\documentclass[10pt,twocolumn,letterpaper]{article}

\usepackage{cvpr}              

\usepackage{graphicx}
\usepackage{amsmath}
\usepackage{amssymb}
\usepackage{booktabs}
\usepackage[pagebackref,breaklinks,colorlinks]{hyperref}

\usepackage{amsmath}
\usepackage{amssymb}
\usepackage{mathtools}
\usepackage{multirow}
\usepackage{enumitem, array}

\usepackage[capitalize]{cleveref}
\crefname{section}{Sec.}{Secs.}
\Crefname{section}{Section}{Sections}
\Crefname{table}{Table}{Tables}
\crefname{table}{Tab.}{Tabs.}

\def\cvprPaperID{4684} 
\def\confName{CVPR}
\def\confYear{2022}

\begin{document}

\title{Cross-domain Few-shot Segmentation with Transductive Fine-tuning}

\author{Yuhang Lu, Xinyi Wu, Zhenyao Wu, Song Wang\\
Univesity of South Carolina
}
\maketitle

\begin{abstract}
Few-shot segmentation (FSS) expects models trained on base classes to work on novel classes with the help of a few support images.
However, when there exists a domain gap between the base and novel classes, the state-of-the-art FSS methods may even fail to segment simple objects.
To improve their performance on unseen domains, we propose to transductively fine-tune the base model on a set of query images under the few-shot setting, where the core idea is to implicitly guide the segmentation of query images using support labels. 
Although different images are not directly comparable, their class-wise prototypes are desired to be aligned in the feature space. 
By aligning query and support prototypes with an uncertainty-aware contrastive loss, and using a supervised cross-entropy loss and an unsupervised boundary loss as regularizations, our method could generalize the base model to the target domain without additional labels.
We conduct extensive experiments under various cross-domain settings of natural, remote sensing and medical images.
The results show that our method could consistently and significantly improve the performance of prototypical FSS models in all cross-domain tasks.
Our code will be made publicly available.
\end{abstract}


\section{Introduction}
\begin{figure}
	\centering
	\includegraphics[width=0.9\linewidth]{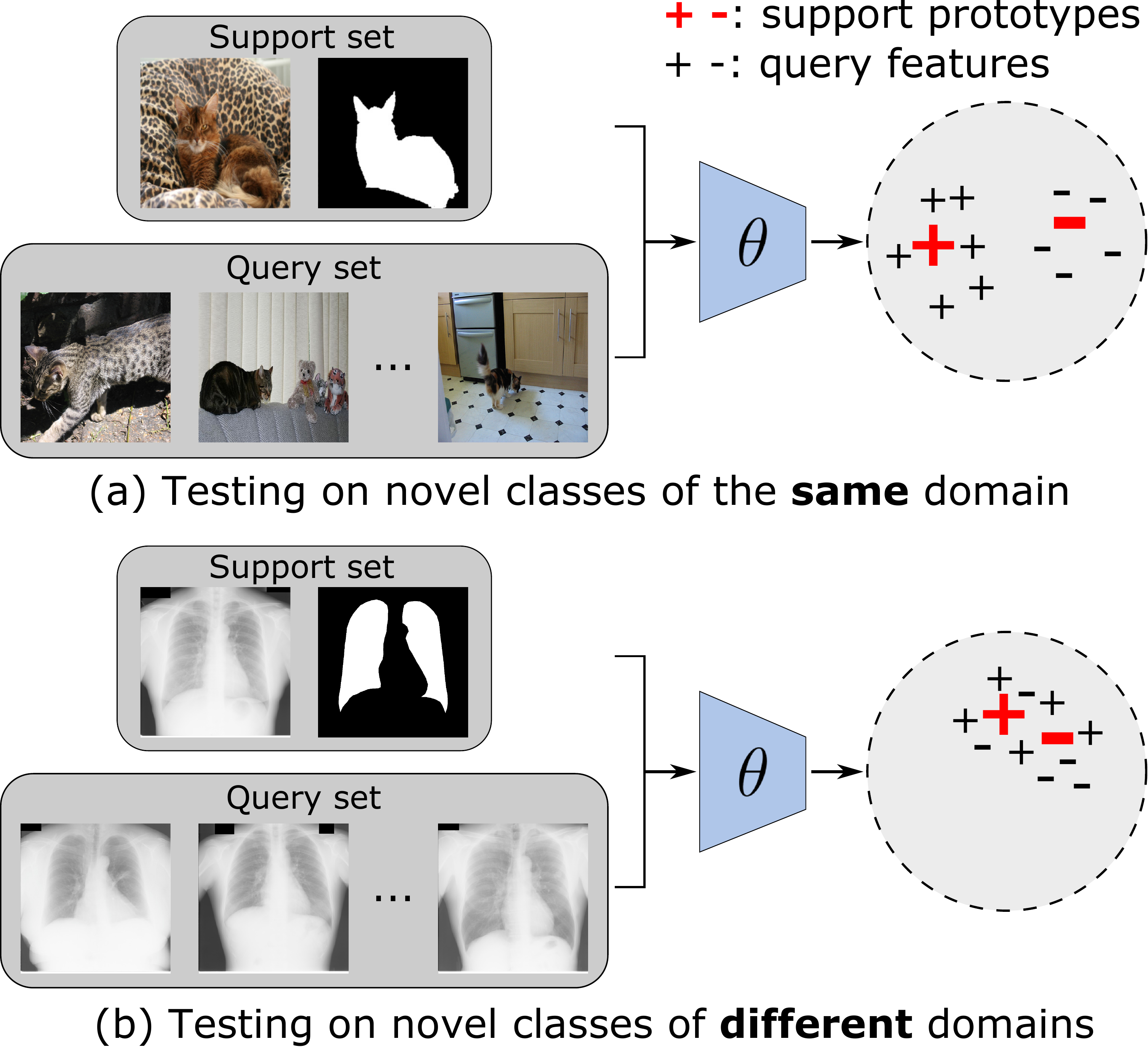}
	\caption{The cross-domain FSS problem. $\theta$ is a prototypical FSS model trained on natural images. When testing on medical images, its discriminability will decreases drastically due to the feature distribution discrepancy. We propose a method to fine-tune $\theta$ on the given support and query sets to bridge the domain gap.}
    \label{fig:intro}
\end{figure}

Few-shot segmentation (FSS) aims to segment unseen object classes in query images by referring to only few labeled (support) images~\cite{shaban2017one}. 
As a potential solution to annotation-efficient segmentation, it has received increasing attention in recent years~\cite{shaban2017one, dong2018few, wang2019panet, zhang2019canet, liu2020part, PFENet, boudiaf2021few, yang2021mining}.
Existing FSS methods typically solve this problem in a meta-learning framework. 
They leverage existing large-scale datasets as base classes, and organize the training data into episodes of query and support images to train a two-branch model that is agnostic to object classes~\cite{boudiaf2021few}.
These methods show good generalization ability in popular FSS benchmarks (\textit{e.g.}, PASCAL-$5^{i}$ and COCO-$20^{i}$) where the base and novel classes are from natural images. 
However, when the novel class comes from different domains, the model trained on base classes often fail due to the large discrepancy of feature distributions~\cite{crossdomainfewshot}, as shown in Fig.~\ref{fig:intro}.
For example, in a real application that motivated this study, we have a set of unlabeled chest X-ray images and need to extract the lungs with one support image.
Although the object to be segmented has relatively low variation across these images, we cannot find any existing FSS model that could extract it very well.
Similar cases can be find in many other fields where the annotations are scarce and require domain expertise to label, such as material images, archaeological images and satellite images.
Therefore, the domain gap greatly limits the application of FSS models.

A straightforward way to address the domain gap is to fine-tune the base model on support images of the novel class.
However, due to the small size of support set in FSS, supervised fine-tuning may lead to overfitting~\cite{liu2018learning}. 
In the field of few-shot classification, transductive fine-tuning has been proven effective to improve the base model's performance on query images of unseen domains~\cite{dhillon2019baseline}. 
They incorporate unlabeled query images into fine-tuning to avoid overfitting.
By minimizing a supervised loss on labeled images and an unsupervised loss on unlabeled image, the base model is optimized on the entire query set.
But transductive fine-tuning for FSS is still under explored.
In the experiment, we show that applying these methods to the cross-domain FSS problem could bring considerable improvement.
However, when proper designs for the segmentation problem are adopted, the cross-domain FSS performance can be further improved.

In this work, we present a transductive fine-tuning method to address the domain gap for prototypical FSS models. 
Instead of learning from labeled and unlabeled images using separate losses, we try to better utilize the connection between query and support images, and the core idea is to \textit{use support labels to implicitly supervise query segmentation}.
Prototypical network is a commonly-used FSS method that generates class-wise prototypes from support images, and makes prediction on query images by finding the closest prototype of each pixel~\cite{snell2017prototypical}.
No matter in the query or support images, pixels from the same class are desired to be closer in the feature space than pixels from other classes. 
Therefore, in absence of query labels, we can use the labels of support images to implicitly guide the query segmentation by aligning their class-wise prototypes, thus making full use of known query and support images.

Given a set of query and support images, our goal is to fine-tune the base FSS model to generalize it to the target domain. 
We first extract image features using the base model, and then generate fine-grained prototypes by clustering in the feature space. 
A prototype contrastive loss is proposed to provide implicit supervision to query images by contrasting its prototypes against support prototypes.
To avoid inaccurate query prototypes, an uncertainty factor is introduced to adjust the loss weight dynamically.
Moreover, our optimization objectives also include a supervised cross-entropy loss that take advantage of support labels, and an unsupervised boundary loss that penalizes scattered query masks.
With these regularizations, our semi-supervised fine-tuning strategy is more stable and more effective. 

In summary, the contributions of this work include:
\begin{enumerate}[topsep=0mm]
\itemsep 0mm
    \item We raise a novel problem of fine-tuning meta-learned models for cross-domain FSS. 
    \item We propose a method to address the domain gap for prototypical FSS models by naturally incorporating unlabeled images into fine-tuning, thus extending their application range from in-domain to cross-domain.
    \item We investigate the performance of representative FSS methods under various cross-domain settings; compare the performance of different fine-tuning methods in cross-domain FSS; and validate the effectiveness of our method in fine-tuning various prototypical models.
\end{enumerate}


\section{Related Work}
\paragraph{Few-shot segmentation}
Most existing FSS methods follow the meta-learning paradigm to train a class-agnostic two-branch model that could leverage the knowledge in support images to perform segmentation. 
Based on how the support knowledge is incorporated, these methods can be categorized into relation- and prototype-based methods. 

Relation-based methods use the two-branch network to learn to compare query images against support images by fusing their features~\cite{sung2018learning}.
For example, Zhang \textit{et al.}~\cite{zhang2019canet} adopt an attention-based fusion scheme to fuse query and support features and predict the query mask using an iterative optimization module.
Liu \textit{et al.}~\cite{liu2020crnet} propose a cross-reference network where two branches concurrently make predictions for both query and support images.
Tian \textit{et al.}~\cite{PFENet} propose to adaptively refine the concatenated query and support features with intra-source enrichment and inter-source interaction. 

Instead of fusing query and support features, prototype-based methods predict masks by computing distance from query features to support prototypes~\cite{snell2017prototypical}.
Dong \textit{et al.}~\cite{dong2018few} made the first attempt to build a prototypical framework for few-shot semantic segmentation. 
Wang \textit{et al.}~\cite{wang2019panet} propose PANet which uses a prototype alignment regularization to maintain a cycle constraint between support and query. 
PPNet proposed by Liu \textit{et al.}~\cite{liu2020part} introduces clustering into FSS, of which the core idea is to decompose the holistic prototype into a set of part-aware prototypes.
Our fine-tuning method focuses on prototypical models to take advantage of prototype consistency to impose implicit supervision.

A recent method ReRPI~\cite{boudiaf2021few} conducts FSS without meta-learning. 
They propose a transductive inference method to fine-tune a linear layer of the base model on each test episode, which also follows the pre-training and fine-tuning approach.
Our method is different from ReRPI as our goal is to optimize existing prototypical FSS models over the entire query set.
Moreover, we build connection from query to support images in fine-tuning, but not treating them separately.
We will show the superiority of our method in addressing domain gap by experiments.

\vspace{-4mm}
\paragraph{Cross-domain few-shot classification}
The domain gap issue has been raised in the field of few-shot classification. 
Similar to our observations in FSS, Chen \textit{et al.}~\cite{chen2019closerfewshot} find that the performance of state-of-the-art meta-learning methods drop drastically when domain gap exists, and can be outperformed by simple baseline methods.
In~\cite{guo2020broader}, the authors systematically study the problem of cross-domain few-shot classification, and establish a comprehensive benchmark from performance evaluation.
Tseng \textit{et al.}~\cite{crossdomainfewshot} propose a feature-wise transformation layer to enhance the generalization ability of meta-learning models by simulating various feature distributions under different domains.
In~\cite{dhillon2019baseline}, Dhillon \textit{et al.} develop a transductive fine-tuning baseline for few-shot classification, which outperforms sophisticated state-of-the-art algorithms. 
These progresses in classification indicate that cross-domain FSS is under explored.

\begin{figure*}[htp]
	\centering
	\includegraphics[width=\linewidth]{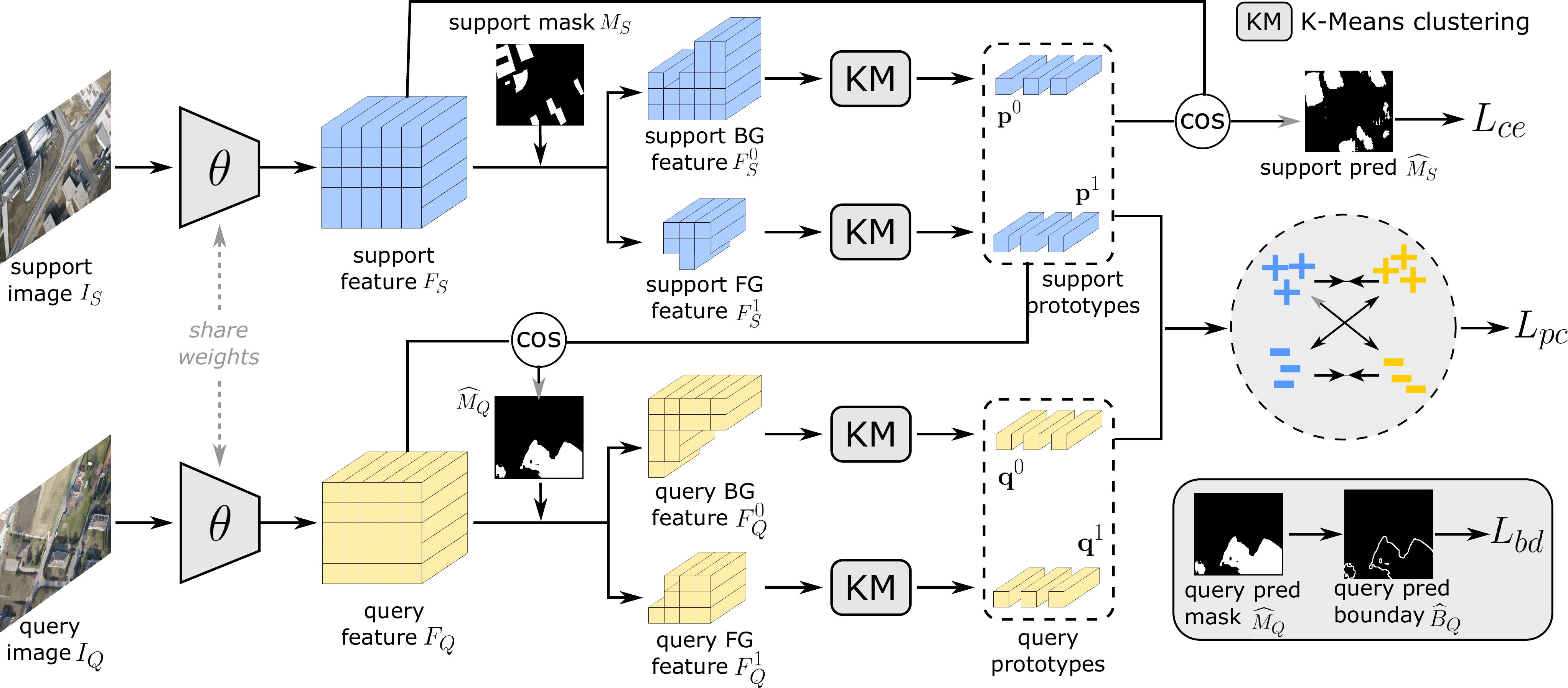}
	\caption{The training pipeline. We illustrate with an example of 1-way 1-shot segmentation. We first generate the fine-grained support prototypes, and then use them to predict the query mask. With the query mask, we generate the query prototypes and compare against the support ones to minimize the prototype contrastive loss $L_{pc}$. Besides, the support GT mask is employed to regularize the training through a cross-entropy loss $L_{ce}$. Finally, a boundary length loss $L_{bd}$ is employed to penalize irregular regions in the query mask.}
    \label{fig:pipeline}
\end{figure*}

\vspace{-4mm}
\paragraph{Semi-supervised segmentation}
Learning from both labeled and unlabeled images is also related to the topic of semi-supervised segmentation. 
Consistency regularization is well-studied for this problem~\cite{mittal2019semi,ouali2020semi}. 
It enforces the consistency of the predictions or intermediate features of various perturbations of the same image. 
Pseudo-labeling is another commonly used approach~\cite{chen2021semi,zou2020pseudoseg}. 
The idea is to utilize the predictions with high confidence of unlabeled images as pseudo ground truth to retrain the model in a iterative manner.
However, these methods have different problem settings with us, where they aim at training new models with a partially labeled dataset, while our goal is to fine-tune existing meta-learning models under the few-shot setting.

\section{Problem setup}
The goal of FSS is to learn to segment any novel classes from only a few support examples. 
In the term $n$-way $k$-shot segmentation, $n$ refers to the number of novel classes (expect for the background), and $k$ refers to the number of support examples.
We denote the support set as $\mathbf{S}=\left \{ S_1, S_2,..., S_k \right \}$, where each sample $S$ consists of a support image $I_S$ and its ground-truth segmentation mask $M_S$. 
Similarly, the query set is denoted as $\mathbf{Q}=\left \{ Q_1, Q_2,... \right \}$, but the query mask $M_Q$ is only available in training. 
The base classes for training and the novel classes for testing are denoted as $C_{base}$ and $C_{novel}$ respectively, where $C_{base} \cap  C_{novel} = \O$.
Meta-learning methods usually adopt episodic training to train class-agnostic FSS models.
They structure the data of $C_{base}$ into episodes, where each episode contains a query image and a support set, thus simulating the testing environment.

When $C_{base}$ and $C_{novel}$ are from different domains, the class-agnostic base model may lose generalization ability due to the feature distribution discrepancy. 
In this work, we address this problem for prototypical FSS models.
Given a prototypical model $\theta$ trained on $C_{base}$, our goal is to generalize $\theta$ to the domain of $C_{novel}$ without additional labels.
We assume the support set $\mathbf{S}$ and the whole query set $\mathbf{Q}$ are available in testing, and fine-tune $\theta$ on $\mathbf{S}$ and $\mathbf{Q}$, which is also referred as transductive fine-tuning.

\section{Method}
We propose Cross-image Prototype Contrast (CPC) to fine-tune prototypical models on the query and support images of unseen domains. 
To avoid overfitting on support images, our core idea is to incorporate unlabeled query images into training by implicitly guiding their segmentation with support prototypes. 
The fine-tuning also follows the episodic training scheme.
As illustrated in Fig.~\ref{fig:pipeline}, we use two branches to process support and query images respectively.
The support branch extracts the prototypes from support images, and outputs a supervised cross-entropy loss.
Without any labels, the query branch leverages support prototypes to predict a query mask, and minimizes an unsupervised boundary loss.
More importantly, a semi-supervised prototype contrast loss is proposed to build connection between two branches by contrasting query prototypes with support prototypes.
Note that our method does not modify the internal structure of $\theta$, so it can be applied to any prototypical encoders.
The following describes each component of CPC in details.



\subsection{Prototype generation}
In each episodic fine-tuning step, the encoder $\theta$ takes a random query image $I_Q$ and the support set $S$ as input.
We first extract fine-grained prototype features on the query and support images.

Given a query image $I_Q$ and a support image $I_S$, we use $\theta$ to obtain their feature maps $F_Q$ and $F_S$, where $F_Q=\theta(I_Q)$, $F_S=\theta(I_S)$.
Subsequently, the support feature $F_S$ can be split into several class-wise feature sets $\left \{ F_{S}^{0}, F_{S}^{1},...,F_{S}^{n} \right \}$ according to the support mask $M_S$, where
\begin{equation}
F_{S}^{i} = \left \{ F_{S}(x,y) | M_S(x,y)=i \right \}.
\label{eq:fs}
\end{equation}
For $k$-shot segmentation, we just put all features of the same class in the same feature set to compute prototypes.

For each class, we can obtain the holistic prototype by averaging all features in the corresponding feature set.
But considering the intra-class diversity of objects (\textit{e.g.},different types of background), the holistic prototype may be too coarse to represent the whole class. 
We follow~\cite{liu2020part} to utilize K-Means clustering to obtain fine-grained prototypes.
For each class $i$, the feature set $F_{S}^{i}$ is split into $c$ clusters, and the cluster centers are denoted by $\left \{ G_{1}^{i}, G_{2}^{i},...,G_{c}^{i} \right \}$.
Finally, we obtain $c$ prototypes for each class, and each prototype is the weighted sum of a cluster center and the holistic prototype, namely
\begin{equation}
\mathbf{p}_j^i = G_j^i + \frac{\lambda }{\left | F_S^i \right |}\sum F_S^i ,
\label{eq:prototype}
\end{equation}
where $i \in [0, n]$, $j \in [1, c]$ and $\lambda$ is a weight set as 0.5.

With the set of support prototypes, we are able to predict the segmentation mask of the query image $I_Q$. 
For each pixel $(x,y)$ in $I_Q$, its class is predicted as the class of the support prototype that has the highest cosine similarity with its feature. 
The class-wise probability map of $I_Q$ is:
\begin{equation}
\widehat{P}_Q^i(x,y) = softmax ( \max_{j \in [1, c]} \left [ {\cos(\mathbf{p}_j^i, F_{Q}(x,y))} \right ]),
\label{eq:predict}
\end{equation}
and the predicted query mask is $\widehat{M}_Q$ can be obtained by: $\widehat{M}_Q(x,y) = \operatorname*{arg\,max}_{i \in [0, n]} \widehat{P}_Q^i(x,y)$.

With the predicted mask $\widehat{M}_Q$ and the query feature map $F_Q$, we follow Eq.(\ref{eq:fs}) and Eq.(\ref{eq:prototype}) to generate the fine-grained prototypes $\mathbf{q}$ of query images, which are pseudo prototypes that are supposed to be aligned with $\mathbf{p}$.

Next, we introduce the optimization objectives of CPC.

\subsection{Prototype contrastive loss}
The role of the prototype contrastive loss is to contrast the support prototypes $\mathbf{p}$ with the query prototypes $\mathbf{q}$. 
For prototypical inference, an ideal feature encoder should map pixels to a feature space where intra-class distances are smaller than inter-class distances. 
When testing on unseen domains, the feature encoder may not be able to extract meaningful features, thus making different classes inseparable in the feature space. 
Therefore, the prototype contrastive loss is designed to improve the discriminability of $\theta$ by reducing the intra-class distance and enlarging the inter-class distance from $\mathbf{p}$ to $\mathbf{q}$.

Given the support prototype $\mathbf{p}$ and the query prototype $\mathbf{q}$, the distance between the $i$-th class of $\mathbf{p}$ and the $j$-th class of $\mathbf{q}$ is computed by:
\begin{equation}
d(\mathbf{p}^i, \mathbf{q}^j) = 1 - \frac{1}{c}\sum_{g=1}^{c} \max_{h \in [1,c]} \cos (\mathbf{p}_h^i, \mathbf{q}_g^j).
\label{eq:d}
\end{equation}
Eq.~(\ref{eq:d}) means that, for each prototype in $\mathbf{p}^i$, we compare it with only the most similar prototype in $\mathbf{q}^j$. 

For the class $j$, the intra-class distance between the query and support prototypes is the distance between $\mathbf{p}^j$ and $\mathbf{q}^j$, and the inter-class distance is averaged over all classes in $\mathbf{p}$ except for $j$:
\begin{equation}
\begin{dcases}
d_{intra}^{j}=d(\mathbf{p}^j, \mathbf{q}^j) \\
d_{inter}^{j}=\frac{1}{n} \sum_{i=0 \wedge i\neq j}^{n} d(\mathbf{p}^i, \mathbf{q}^j).
\label{eq:intra_inter}
\end{dcases}
\end{equation}

Solely minimizing $d_{intra}$ or maximizing $d_{inter}$ will lead to collapse solutions.
We employ the triplet loss to optimize them simultaneously. 
To avoid overly maximizing the inter-class distance to make $\mathbf{p}$ and $\mathbf{q}$ negatively correlated, we use a margin value $m$ to balance the intra- and inter-class distances, $m=0.2$.
The prototype contrast loss is:
\begin{equation}
{L}_{pc} = \sum_{j=0}^{n} \max (d_{intra}^{j} - d_{inter}^{j} + m, 0).
\label{eq:tri_loss}
\end{equation}

As a semi-supervised loss, $L_{pc}$ takes advantage of unlabeled images to enrich training data. 
But it is not as reliable as supervised losses. 
When aligning query and support prototypes, we are actually assuming the query mask that produces the query prototype is accurate. 
However, it is not this case in practice, especially in the beginning of fine-tuning. 
If the query mask is wrong, pushing the query features to the wrong support prototype may affect the fine-tuning adversely.
To alleviate this problem, we propose to weight $L_{pc}$ by a \textbf{dynamic uncertainty} during fine-tuning.

Given the probability map $\widehat{P}_Q$ of a query image, we compute its global uncertainty $w_{un}$ as the average ratio of the second largest probability to the largest probability:
\begin{equation}
w_{un}=\frac{1}{\left | \Phi \right |}\sum_{(x,y)\in \Phi }\frac{\max_{i \in [0,n]\wedge i\neq j} \widehat{P}_Q^i(x,y)}{\max_{i \in [0,n]} \widehat{P}_Q^i(x,y)},
\label{eq:uncertainty}
\end{equation}
where $j$ is the class index with the largest probability, and $\Phi$ is the set of all pixels in $\widehat{P}_Q$.
It indicates how confident the encoder is on the predicted query mask.
We use $w_{un}$ to dynamically adjust the weight of the semi-supervised loss.

\subsection{Supervised cross-entropy loss}
To make full use of support labels, we employ a cross-entropy loss to supervise the prediction on support images. 
Similar to Eq.(\ref{eq:predict}), we use the support prototypes $\mathbf{p}$ to predict the softmax probability map $\widehat{P}_S$ on $F_S$.
The cross-entropy loss of the support image is calculated as:
\begin{equation}
L_{ce}=-\frac{1}{(n+1)\left | \Psi  \right |}\sum_{i=0}^{n}\sum_{(x,y)\in \Psi } M_S^i(x,y) \cdot log(\widehat{P}_S^i(x,y)),
\label{eq:ce_loss}
\end{equation}
where $\Psi$ is the set of valid pixels in $M_S$.

$L_{ce}$ forces $\theta$ to extract similar features for pixels of the same class in $I_S$.
Using $L_{ce}$ alone will lead to overfitting, especially for 1-shot segmentation, but when using with $L_{pc}$ together, it provides strong regularization to prevent the optimization from deviating from the right direction.

\setlength{\tabcolsep}{2mm}
\begin{table*}[]
\centering
\caption{The IoU (\%) results of using our method to fine-tune various prototypical models on four cross-domain FSS tasks.}
\label{table:cross-domain}
\begin{tabular}{l|ccccc|ccccc}
\toprule
\multirow{2}{*}{Method} & \multicolumn{5}{c|}{1-shot}          & \multicolumn{5}{c}{5-shot}         \\ & Rooftop & Road & Lung & Knee & Mean & Rooftop & Road & Lung & Knee & Mean \\
\midrule
ImageNet weights                   & 19.5    & 25.6 & 59.9 & 72.5 & 44.4 & 34.1    & 32.5 & 65.9 & 75.4 & 52.0    \\
ImageNet weights + Ours & \textbf{32.8} & \textbf{30.3} & \textbf{73.7} & \textbf{73.7} & \textbf{52.6} & \textbf{40.4} & \textbf{37.1} & \textbf{78.8} & \textbf{86.1} & \textbf{60.6}     \\
\midrule
PANet~\cite{wang2019panet} & 27.8    & 31.4 & 65.1 & 72.0 & 49.1 & 37.5    & 34.7 & 65.1 & 73.2 & 52.6     \\
PANet + Ours & \textbf{36.0} & \textbf{33.1} & \textbf{81.7} & \textbf{84.9} & \textbf{58.9} & \textbf{53.5} & \textbf{42.2} & \textbf{85.8} & \textbf{88.3} & \textbf{67.5}     \\ 
\midrule
PPNet~\cite{liu2020part} & 22.3    & 31.2 & 61.6 & 73.8 & 47.2 & 39.4    & 34.9 & 67.8 & 75.3 & 54.4     \\
PPNet + Ours   & \textbf{34.5}        & \textbf{39.7}        & \textbf{81.6}            & \textbf{83.9}            & \multicolumn{1}{c|}{\textbf{59.9}} & \textbf{55.9}        & \textbf{46.7}        & \textbf{88.9}        & \textbf{94.8}        & \textbf{71.6}     \\
\midrule
PFENet~\cite{PFENet} & 16.8          & 1.3           & 41.6          & 55.6          & 28.8          & 20.5          & 1.9           & 39.7          & 56.5          & 29.7  \\
\midrule
ReRPI~\cite{boudiaf2021few} & 5.6  & 8.7  & 65.6 & 76.8 & 39.2 & 27.4 & 18.6 & 67.8 & 79.6 & 48.4      \\
\bottomrule
\end{tabular}
\end{table*}

\subsection{Unsupervised boundary loss} 
Because of the pixel-wise classification strategy, prototypical models may produce ``small islands'' in the prediction mask at some ambiguous pixels, which will downgrade the visual quality of results. 
Therefore, we employ an unsupervised boundary loss~\cite{chen2019learning} to penalize the boundary length in the query prediction $\widehat{M}_Q$, which is formulated as:
\begin{equation}
L_{bd}=\sum_{(x,y)\in M_Q}^{}\sqrt{\left | (\triangledown_{M_Q}^\mathbf{x}(x,y))^2 + (\triangledown_{M_Q}^\mathbf{y}(x,y))^2 \right |},
\label{eq:bd_loss}
\end{equation}
where $\triangledown^\mathbf{x}$ and $\triangledown^\mathbf{y}$ are the gradient of $M_Q$ in the $\mathbf{x}$ and $\mathbf{y}$ directions, respectively.

Finally, the optimization objective of CPC is written as:
\begin{equation}
L= L_{ce} + (1-w_{un})L_{pc} + L_{bd}.
\label{eq:final_loss}
\end{equation}

\section{Experiments}

\subsection{Experimental setup}




\paragraph{Datasets}
To evaluate the performance of our method, we resort to four public datasets from different domains, including two remote sensing and two medical image datasets:
1) \textbf{Rooftop}: Aerial images from the EPFL Rooftop dataset~\cite{rooftop};
2) \textbf{Road}: Satellite images for the EPFL Road Segmentation Challenge~\footnote{\url{https://www.crowdai.org/challenges/epfl-ml-road-segmentation}};
3) \textbf{Lung}: Chest X-ray images from the SCR dataset~\cite{scr} for lung segmentation;
4) \textbf{Knee}: X-ray images of knees sampled from the OAI dataset~\footnote{\url{https://nda.nih.gov/oai}}~\cite{lu2020contour}.
Each of them has only one foreground class.
Models trained on PASCAL-$5^i$~\cite{shaban2017one} are tested on these four datasets to simulate realistic cross-domain FSS tasks.

\vspace{-4mm}
\paragraph{Implementation details}
Our method is implemented in PyTorch~\cite{paszke2019pytorch}. We fine-tune models using the SGD optimizer with a fixed learning rate of $1\times10^{-5}$ and a momentum of 0.9 for 1000 iterations. All images are resized to $417 \times 417$ in both training and testing. 
As a transductive method, we need to fix support images before fine-tuning.
We select the most representative images in each dataset as the support images, and the rest are query images.
We use a ResNet-50~\cite{he2016deep} encoder pretrained on ImageNet~\cite{russakovsky2015imagenet} to extract all image features, and split them into $k$ clusters by K-Means, then the cluster centers are selected as support images.
For fair comparison, all methods use the same support images.

\subsection{Performance evaluation}

\begin{figure}[]
	\centering
	\includegraphics[width=\linewidth]{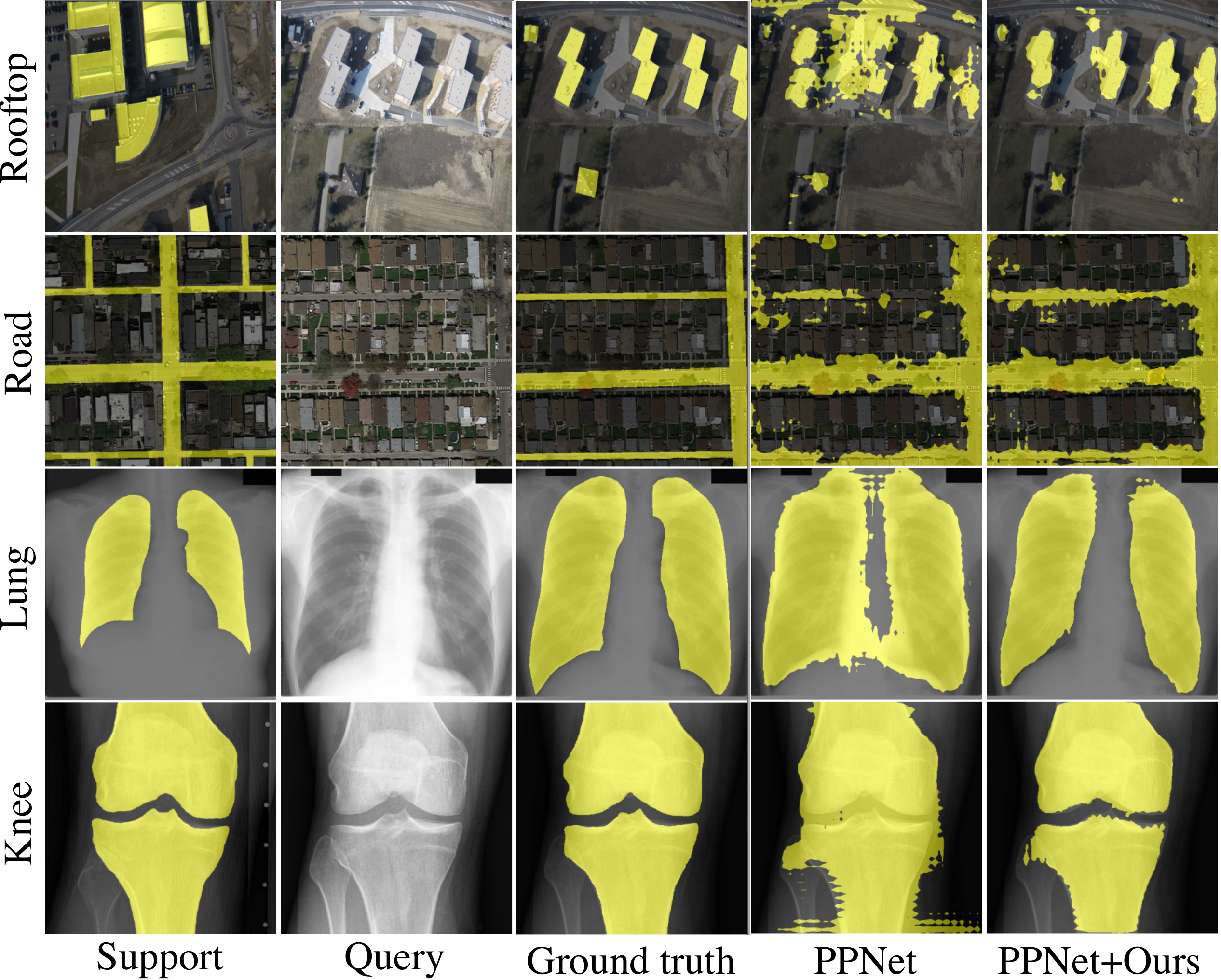}
	\caption{Qualitative results. By comparing the last two columns in this figure, we can observe the improvement brings by fine-tuning using our method for PPNet.}
    \label{fig:res_vis}
\end{figure}

\setlength{\tabcolsep}{2mm}
\begin{table*}[]
\centering
\caption{The IoU (\%) results of using three different methods to fine-tune PANet and PPNet on four cross-domain FSS tasks.}
\label{table:cmp}
\begin{tabular}{l|ccccc|ccccc}
\toprule
\multirow{2}{*}{Method} & \multicolumn{5}{c|}{1-shot}          & \multicolumn{5}{c}{5-shot}         \\ & Rooftop & Road & Lung & Knee & Mean & Rooftop & Road & Lung & Knee & Mean \\
\midrule
PANet~\cite{wang2019panet} & 27.8                 & 31.4                 & 65.1                 & 72.0                 & 49.1                               & 37.5                 & 34.7                 & 65.1                 & 73.2                 & 52.6     \\
PANet + Sup-FT~\cite{liu2020crnet}
 & 32.0                 & 32.9                 & 79.5                 & 82.6                 & 56.8                               & 47.3                 & 39.6                 & 85.6                 & 89.3                 & 65.5     \\
PANet + Trans-FT~\cite{dhillon2019baseline} & 30.2                 & 28.1                 & 78.9                 & 81.7                 & 54.7                               & 48.4                 & 39.5                 & 85.2                 & 88.2                 & 65.3     \\
PANet + Ours & \textbf{36.0}        & \textbf{33.1}        & \textbf{81.7}        & \textbf{84.9}        & \textbf{58.9}                      & \textbf{53.5}        & \textbf{42.2}        & \textbf{85.8}        & \textbf{91.1}                 & \textbf{68.2}    \\
\midrule
PPNet~\cite{liu2020part} & 22.3                 & 31.2                 & 61.6                 & 73.8                 & 47.2                               & 39.4                 & 34.9                 & 67.8                 & 75.3                 & 54.4     \\
PPNet + Sup-FT~\cite{liu2020crnet} & 28.9                 & 33.2                 & 73.1                 & 81.5                 & 54.2                               & 49.5                 & 40.4                 & 87.7                 & 88.1                 & 66.4     \\
PPNet + Trans-FT~\cite{dhillon2019baseline} & 17.5                 & 26.2                 & 74.2                 & 83.6                 & 50.4                               & 48.5                 & 35.9                 & 86.1                 & 86.2                 & 64.2     \\
PPNet + Ours & \textbf{34.5}        & \textbf{39.7}        & \textbf{81.6}        & \textbf{83.9}        & \textbf{59.9}                      & \textbf{55.9}        & \textbf{46.7}        & \textbf{88.9}        & \textbf{94.8}        & \textbf{71.6}    \\
\bottomrule
\end{tabular}
\end{table*}

\subsubsection{Evaluation on cross-domain FSS tasks}
We first evaluate the performance of the proposed fine-tuning method on cross-domain FSS tasks. 
We fine-tune the models of two representative prototypical methods, PANet~\cite{wang2019panet} and PPNet~\cite{liu2020part}. 
We also directly fine-tune the ResNet-50 pretrained on ImageNet, which is the initial weight of many FSS models.
It tries to skip meta-learning to directly deploy ImageNet weights to downstream FSS tasks.
Without any fine-tuning, we also evaluate a relation-based method PFENet~\cite{PFENet} and a transductive inference method ReRPI~\cite{boudiaf2021few} on cross-domain tasks.
Except for the ImageNet weights, all models are pretrained on the fold-0 of PASCAL-$5^i$ with a ResNet-50 backbone.

The evaluation results are reported in Table~\ref{table:cross-domain}. 
We can see that our method significantly improve the performances of all three prototypical models on cross-domain FSS. 
Among them, we achieve the greatest improvement on PPNet with 12.7\% and 17.2\% higher IoU, averaged on four tasks, for 1-shot and 5-shot segmentation, respectively. 
The gain on medical images is more significant than it on remote sensing images, because objects with less appearance variation are more suitable for few-shot learning once the feature distribution shift is removed.
Directly fine-tuning on ImageNet weights has inferior performance to fine-tuning on meta-learned models. 
It indicates that pretraining on large-scale base classes is beneficial.
Besides, we notice that prototype-based methods have better generalization ability to large domain gap than the representative relation-based and transductive inference methods.
We show the results of four example images in Fig.~\ref{fig:res_vis}.



\setlength{\tabcolsep}{1.8mm}
\begin{table*}
\caption{The IoU (\%) results of using our method to fine-tune PPNet on two in-domain FSS datasets - PASCAL-$5^i$ and COCO-20$^i$.}
\label{table:in-domain}
\centering
\begin{tabular}{l|l|ccccc|ccccc}
\toprule
\multirow{2}{*}{Dataset} & \multirow{2}{*}{Method} & \multicolumn{5}{c|}{1-Shot}              & \multicolumn{5}{c}{5-Shot}               \\
                        &                           & Fold-0 & Fold-1 & Fold-2 & Fold-3 & Mean & Fold-0 & Fold-1 & Fold-2 & Fold-3 & Mean \\
\midrule
\multirow{2}{*}{PASCAL-5$^i$}        & PPNet~\cite{liu2020part}                     & 59.8 & 65.9 & 64.4 & 58.5 & 62.1 & 70.2 & 71.1 & 71.0 & 60.7 & 68.2 \\
    & PPNet + Ours                     & \textbf{67.3} & \textbf{67.3} & \textbf{68.0} & \textbf{58.6} & \textbf{65.3} & \textbf{72.3} & \textbf{73.1} & \textbf{73.3} & \textbf{64.4} & \textbf{70.8} \\ 
\midrule
\multirow{2}{*}{COCO-20$^i$}        & PPNet~\cite{liu2020part}                     & 39.9 & \textbf{41.5} & 45.1 & 38.1 & 41.1 & 49.2 & 52.6 & 48.7 & \textbf{46.3} & 49.2 \\
    & PPNet + Ours                     & \textbf{41.8} & 41.2 & \textbf{45.7} & \textbf{39.4} & \textbf{42.0} & \textbf{50.3} & \textbf{53.9} & \textbf{49.5} & 45.7 & \textbf{49.8} \\ 
\bottomrule
\end{tabular}
\end{table*}

\subsubsection{Comparison with other fine-tuning methods}
Furthermore, we compare with two existing fine-tuning strategies to demonstrate the superiority of our method. 
The first approach is fine-tuning with only support images in a supervised manner.
We follow ~\cite{liu2020crnet} to structure $k$ support images into $k^2$ pairs of support-query episode, and then perform fully-supervised episodic training.
The second approach is a transductive fine-tuning method proposed for few-shot classification~\cite{dhillon2019baseline}.
Besides the cross-entropy loss on support images, they use an unsupervised loss to minimize the entropy of predictions on query images.
We refer to these two approaches as ``Sup-FT'' and ``Trans-FT'', respectively.

The results of using all methods to fine-tune PANet and PPNet are reported in Table~\ref{table:cmp}. 
From Table~\ref{table:cmp}, we can see that: 
1) Our method has better performance than the other two fine-tuning methods. We averagely outperform Sup-FT, the better of the two, by 5.7\% of IoU on 1-shot and 5.2\% on 5-shot, respectively;
2) When using our improvement as reference, Sup-FT and Trans-FT have inferior performance on rooftop and road than on lung and knee. In 1-shot segmentation, Trans-FT results in lower IoU on rooftop (22.3\% to 17.5\%) and road (31.2\% to 26.2\%) after fine-tuning.
It indicates that the information of unlabeled images is not well leveraged by directly minimizing the entropy.
3) When using 5 support images, all fine-tuning methods yield better performance, while our method is still the best of three.

\begin{figure*}[h]
	\centering
	\includegraphics{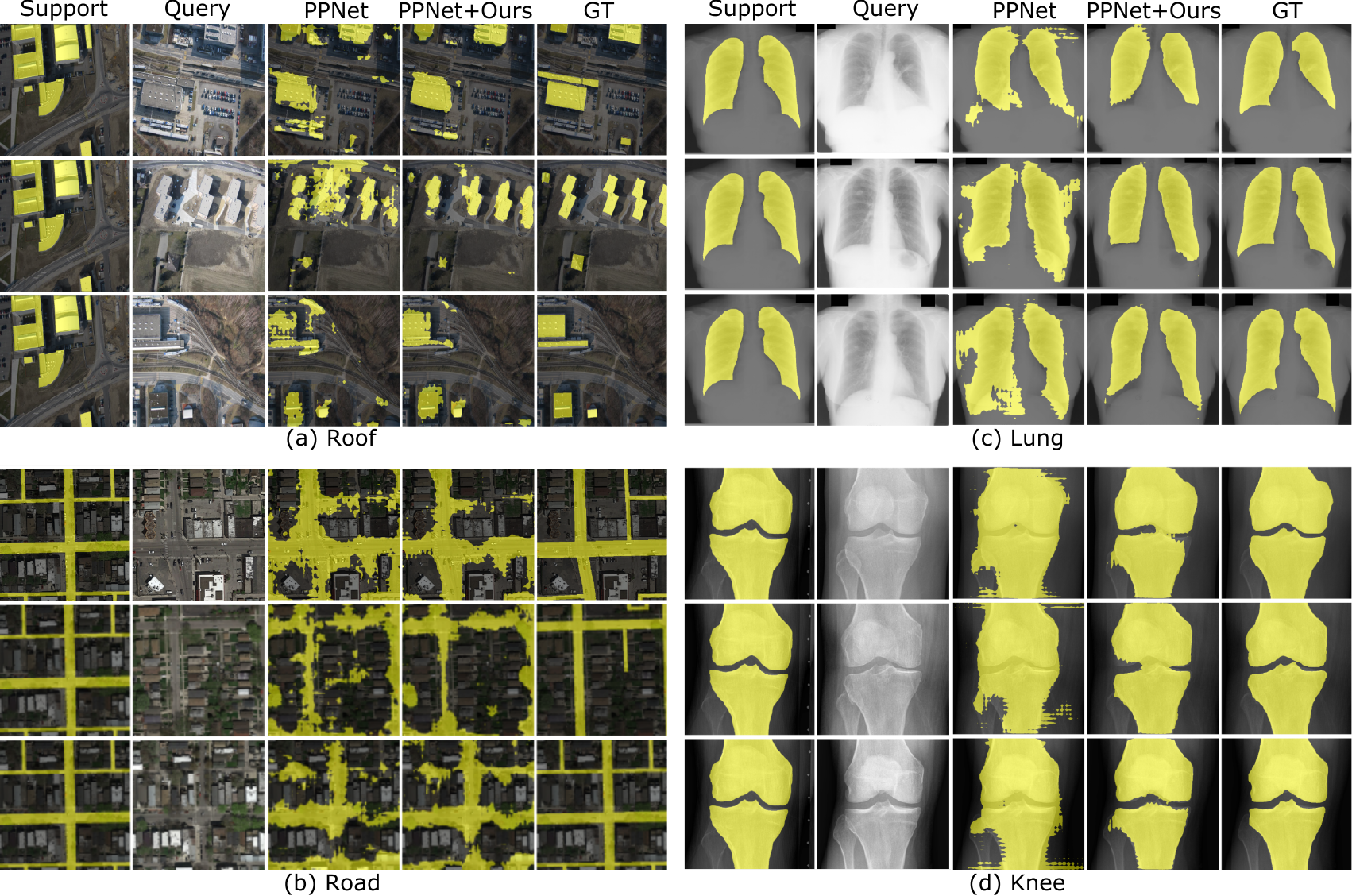}
	\caption{Qualitative results of our method on four cross-domain FSS tasks.}
	\label{fig:more_vis}
\end{figure*}

\begin{figure*}[h]
	\centering
	\includegraphics[width=1\linewidth]{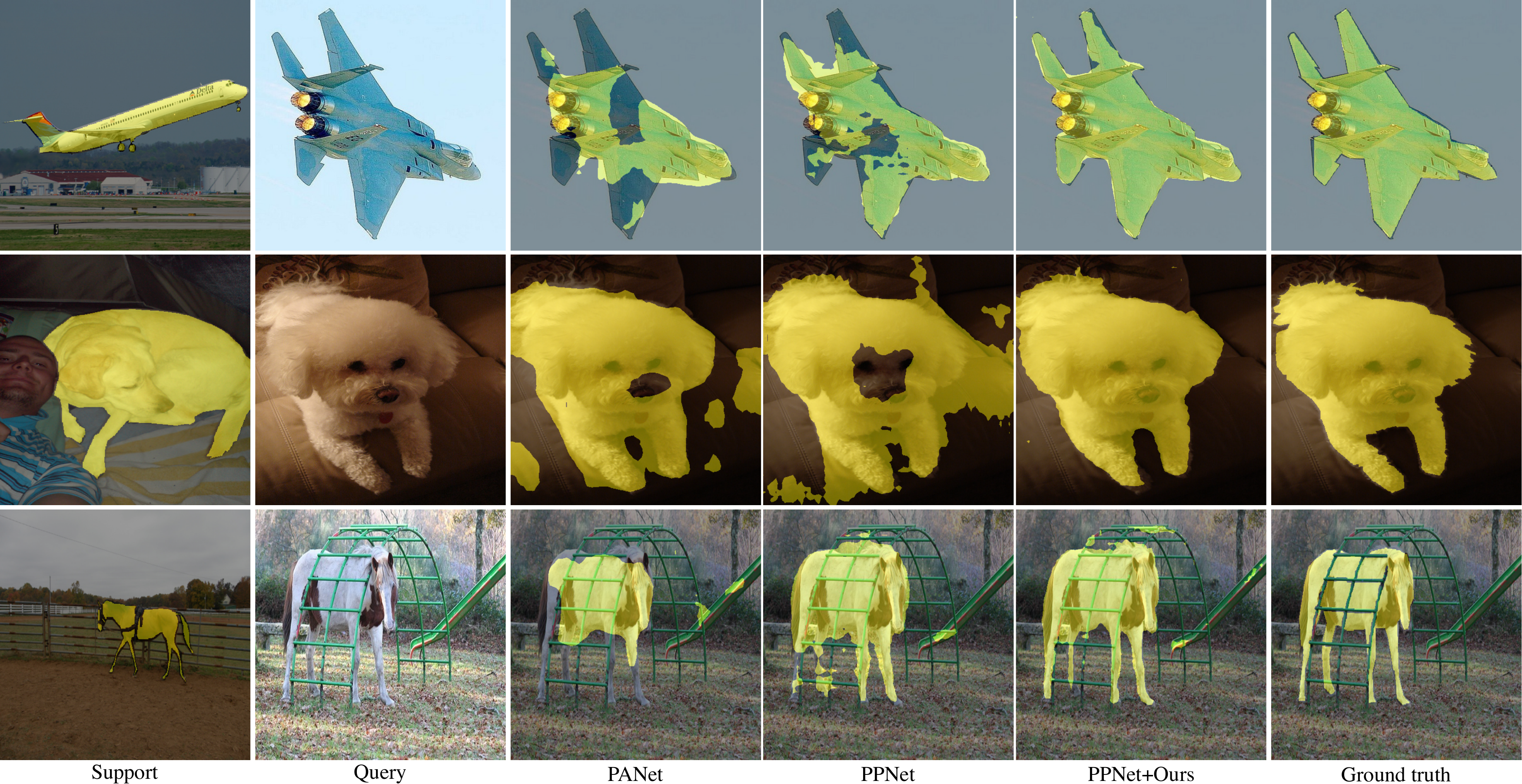}
	\caption{Qualitative results of our method on the in-domain FSS dataset PASCAL-$5^i$.}
	\label{fig:voc}
\end{figure*}

In Fig.~\ref{fig:more_vis} and Fig.~\ref{fig:voc}, we display more qualitative results of our method on the cross-domain and in-domain FSS tasks, respectively.

\subsubsection{Evaluation on in-domain FSS tasks}
When there is no significant domain discrepancy between the base and novel classes, it is still beneficial to use our method to fine-tune the base models.
To validate this point, we conduct two experiments to fine-tune the PPNet~\cite{liu2020part} model on two in-domain FSS datasets PASCAL-5$^i$ and COCO-20$^i$, respectively.
The detailed results are reported in Table~\ref{table:in-domain}.
In Table~\ref{table:in-domain}, we can observe that our method increases the mean IoU of PPNet by 3.2\% and 2.6\% on 1-shot and 5-shot tasks of PASCAL-5$^i$, respectively.
The images in COCO-20$^i$ are more difficult, but our methods still make improvements of 0.9\% and 0.6\% on 1-shot and 5-shot tasks, respectively.
Compared to our performance on cross-domain tasks, the improvements of our method on in-domain tasks are relatively minor, but the results still prove that our method works no matter if there is a domain gap.

\begin{table}
\centering
\caption{The IoU (\%) results of using different combinations of loss terms in Eq.(\ref{eq:final_loss}) to fine-tune PPNet for 1-shot segmentation.}
\label{table:loss}
\begin{tabular}{c|ccccc}
\toprule
  & Rooftop & Road & Lung & Knee & Mean \\
\midrule
w/o $L_{ce}$ & 11.8          & 27.7          & 56.7          & 58.9          & 38.8    \\
w/o $L_{pc}$ & 28.9          & 33.7          & 74.5          & 82.0          & 54.8     \\
w/o $L_{bd}$ & \textbf{35.2} & 38.8          & 80.7          & 79.7          & 58.6    \\ \hline
all           & 34.5          & \textbf{39.7} & \textbf{81.6} & \textbf{83.9} & \textbf{59.9}    \\
\bottomrule
\end{tabular}
\end{table}

\begin{figure}[t]
	\centering
	\includegraphics[width=\linewidth]{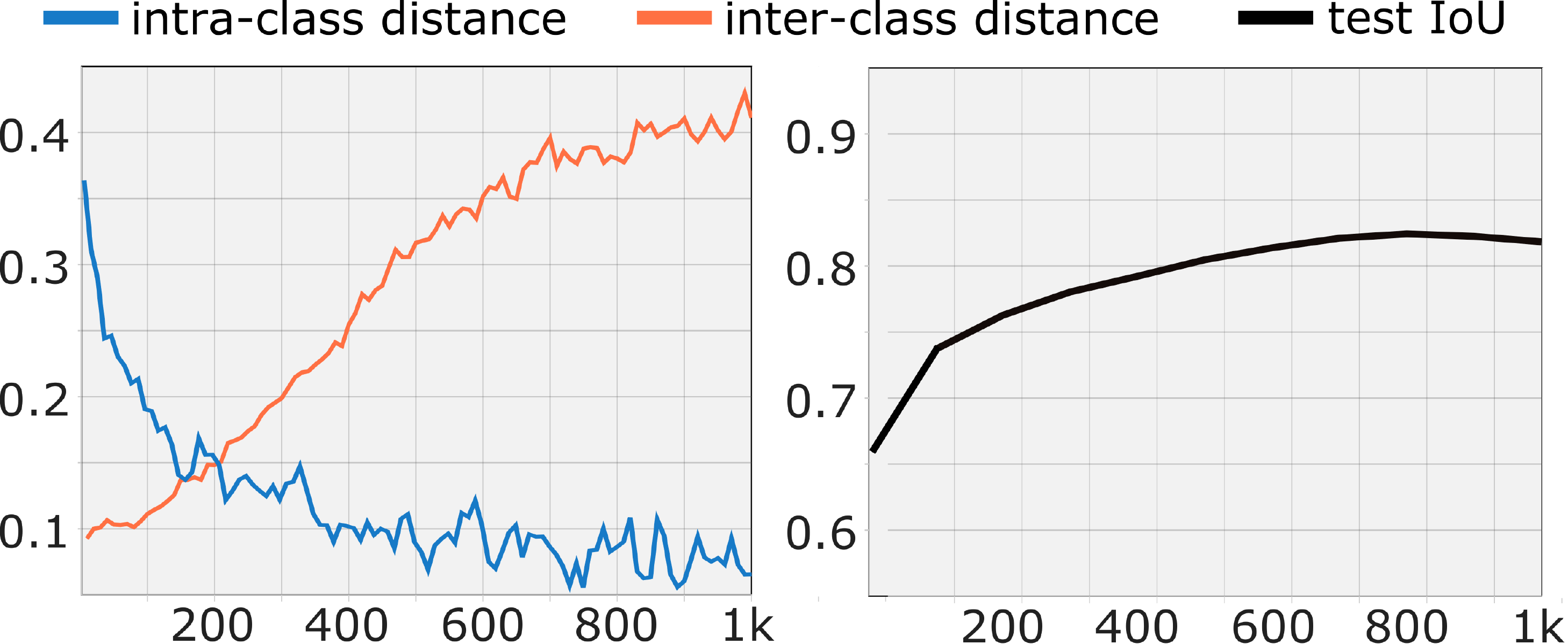}
	\caption{The curves of intra-, inter-class prototype distance, and test IoU v.s. fine-tuning steps.}
    \label{fig:loss}
\end{figure}

\subsection{Ablation studies}
\subsubsection{Impact of each loss term}
In Eq.(\ref{eq:final_loss}), we use three losses to optimize the fine-tuned model, respectively are the supervised cross-entropy loss $L_{ce}$, the prototype contrastive loss $L_{pc}$, and the unsupervised boundary loss $L_{bd}$. 
In this experiment, we remove one of three terms each time and observe how the performance changes.
The IoU results of using modified losses to fine-tune PPNet for 1-shot segmentation on four tasks are reported in Table~\ref{table:loss}.
The performance drops bring by removing $L_{ce}$, $L_{pc}$ and $L_{bd}$ are 21.1\%, 5.1\% and 1.3\%, respectively. 
It indicates the supervised loss $L_{ce}$ is the most essential term in Eq.(\ref{eq:final_loss}).
Without $L_{ce}$ as regularization, the semi-supervised and unsupervised losses may not be able to find the right direction for optimization.
But removing $L_{pc}$ also results in substantial decrease, it shows the effectiveness of support prototypes in guiding query segmentation.

To better understand the behavior of $L_{pc}$, we visualize how the intra- and inter-class prototype distances change during fine-tuning in Fig.~\ref{fig:loss}. We can see that, with the push and pull forces provided by $L_{pc}$, the intra-class distance keeps decreasing while the inter-class distance keeps increasing. It means the discriminability of the base model is improving, thus resulting higher test IoU.

\subsubsection{Impact of the uncertainty weight}
In Eq.(\ref{eq:uncertainty}), we introduced the dynamic uncertainty to adjust the weight of $L_{pc}$. 
The purpose was to prevent from aligning wrong query prototypes when the query prediction is not accurate. 
In this experiment, we remove this term from the objective function Eq.(\ref{eq:uncertainty}) to validate its usefulness.
The results of our method with and without the uncertainty weight $w_{un}$ are reported in Table~\ref{table:uncertainty}.
We can see that removing $w_{un}$ leads to a 9.8\% IoU decrease on average.
The decreases on medical images are more significant than them on remote sensing images.
This is because the increased weight of $L_{pc}$ diminished the influence of $L_{ce}$ in fine-tuning, which could provide very strong supervision for medical images, even if there is only one support image. 
In contrast, the fine-tuning of remote sensing images relies more on unlabeled images and $L_{pc}$, because one support image cannot cover all types of rooftops and roads.

\begin{table}[]
\centering
\caption{The IoU (\%) results of our method with and without the uncertainty weight in fine-tuning PPNet for 1-shot segmentation.}
\label{table:uncertainty}
\begin{tabular}{c|ccccc}
\toprule
              & Rooftop & Road & Lung & Knee & Mean \\
\midrule
w/o $w_{un}$ & 31.7          & \textbf{39.9}          & 62.1          & 66.5          & 50.1    \\
w/ $w_{un}$           & \textbf{34.5}          & 39.7 & \textbf{81.6} & \textbf{83.9} & \textbf{59.9}    \\
\bottomrule
\end{tabular}
\end{table}

\subsubsection{Changing support images}
Our support images are selected as the most representative images in the deep feature space, \textit{i.e.}, the cluster centers. 
In this experiment, we investigate the performance of our method in fine-tuning PPNet with different support images for 5-shot segmentation.
We set five consecutive random seeds to generate the support lists. 
The experiment results are shown in Table~\ref{table:support}. 
We can see that, with five different sets of random support images, the mean IoU ranges from 67.7\% to 70.1\%, which is relatively stable.
But the selected cluster centers still achieve better average performance than random support images.

\setlength{\tabcolsep}{1.5mm}
\begin{table}[t]
\centering
\caption{The IoU (\%) results of using random support images in fine-tuning PPNet for 5-shot segmentation.}
\label{table:support}
\begin{tabular}{c|ccccc}
\toprule
Random seed & Rooftop & Road & Lung & Knee & Mean \\
\midrule
1000 & 55.2          & 45.2          & 87.3          & 91.2          & 69.7    \\
1001 & 56.7          & 41.6          & 86.4          & 91.8          & 69.1     \\
1002 & \textbf{57.1} & 38.9          & 85.5          & 89.6          & 67.8     \\
1003 & 51.7          & 43.3          & 86.2          & 89.5          & 67.7      \\
1004 & 55.8          & 46.0          & \textbf{89.7} & 88.5          & 70.1       \\
\midrule
cluster centers           & 55.9          & \textbf{46.7} & 88.9          & \textbf{94.8} & \textbf{71.6}    \\
\bottomrule
\end{tabular}
\end{table}

\subsubsection{Testing on unseen images}
Since CPC is a transductive fine-tuning method that directly optimizes on the query set, we want to know that if the fine-tuned model can generalize to images unseen in fine-tuning. 
For this purpose, we evenly split each dataset to two subsets, one for fine-tuning and one for testing.
Images in the test subset are from the same class but unseen in fine-tuning. 
For fair comparison, we perform two-fold cross validation in this experiment. 
The results are aggregated over two folds.
We report results of this experiment in Table~\ref{table:inductive}. 
As we can see, the performances of our method on unseen images is 2.2\% lower than it on seen images on average, but still 10.1\% higher than the base model. 
It indicates that our model is not overfitted on the fine-tuning sets and can used on other images of the novel class.

\setlength{\tabcolsep}{2mm}
\begin{table}[t]
\centering
\caption{The IoU(\%) results on seen and unseen images. We fine-tune PPNet for 1-shot segmentation on the ``seen`` subset of each dataset, and test on the ``unseen'' subset.}
\label{table:inductive}
\begin{tabular}{c|ccccc}
\toprule
Subset & Rooftop & Road & Lung & Knee & Mean \\
\midrule
Seen & \textbf{34.3}          & \textbf{37.8}          & \textbf{82.3}          & \textbf{83.7}          & \textbf{59.5}    \\
Unseen & 30.5          & 36.6          & 80.2          & 81.9          & 57.3    \\ 
\bottomrule
\end{tabular}
\end{table}

\subsection{Visualization of features before and after fine-tuning}
\begin{figure}[h]
	\centering
	\includegraphics[width=1\linewidth]{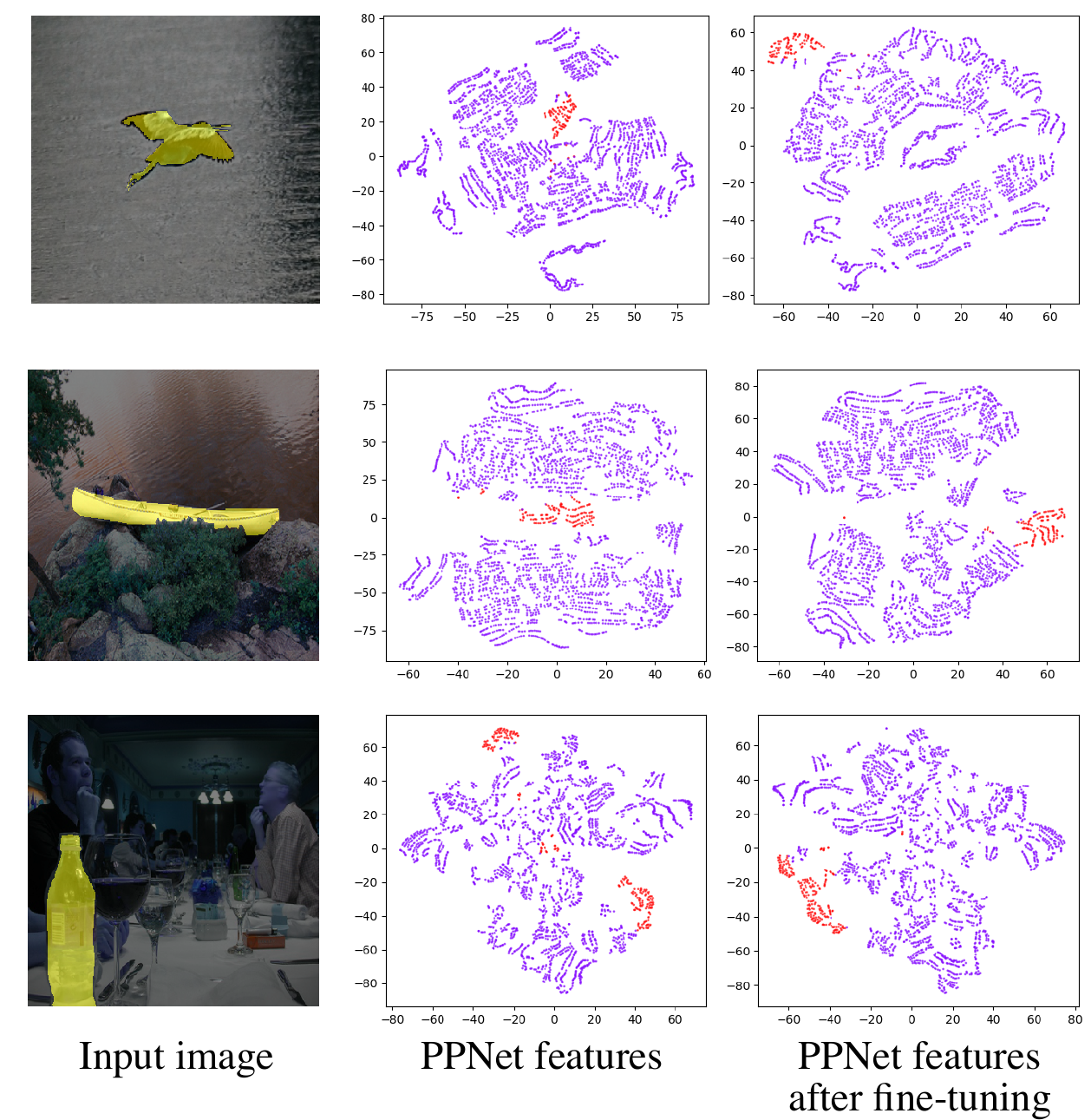}
	\caption{Comparison of the t-SNE features of PPNet before and after using our method for fine-tuning. The red points refer to ground truth foreground pixels and blue points refer to background pixels.}
	\label{fig:tsne}
\end{figure}

To better understand the proposed FSS fine-tuning method, we visualize the image features before and after fine-tuning using t-SNE~\cite{van2008visualizing}. 
In Fig.~\ref{fig:tsne}, we show the feature distribution maps of three example images. 
Specifically, we first use the encoder of the PPNet model or the fine-tuned PPNet model to obtain the features of the input image, and then identify the foreground and background features using the ground truth mask.
Last, we use t-SNE to project the features to a 2D space, as shown in Fig.~\ref{fig:tsne}, and observe the distance between the foreground and background features. 
Ideally, the foreground features should be compact and far away from the background features. 
From Fig.~\ref{fig:tsne}, we can see that the foreground features of the PPNet model are loose or surrounded by background features.
After fine-tuning with our method, the foreground features are separated from the background features.
It further proves the effectiveness of our method in improving the discrimination ability of prototypical FSS models.

\subsection{Discussion}
With extensive experiments and ablation studies, we validated the effectiveness of our method in addressing domain gap for prototypical models. 
But it still has several limitations.
As a transductive method, we expect to fine-tune on a bunch of query images as a whole.
When we have very few query images, it may not be efficient to use our method to fine-tune models. 
On a single Nvidia 1080Ti GPU, our method takes approximately 11 minutes to fine-tune a ResNet-50 encoder for 1-shot segmentation and 25 minutes for 5-shot segmentation.
Moreover, as a common problem of FSS methods, the selection of support images is critical to our performance. 
When unlabeled images are scarce, finding a representative image as support may be difficult.
However, considering the application scenario of FSS, we usually have much more query images than support images in practice.
In this case, fine-tuning the base model using our method is usually beneficial, especially for cross-domain tasks.

\section{Empirical study on support image selection}
In this section, we conduct an empirical study on the impact of support image selection in few-shot segmentation. 
In the test set of each class in PASCAL-$5^i$, we alternately use every image as support, and test the performance of PPNet on this class under the 1-way 1-shot setting.
For example, the test set of the ``areoplane'' class contains 81 images. 
Each time, we use 1 image as the support and test on other 80 images using the PPNet model, and take the mean IoU on 80 images as the performance of the selected support image. 
After trying out all 81 images, we find the minimum and maximum IoU, and compute the IoU averaged on 81 images.
Finally, we compare these results with the performance of the representative support image, which is selected as the cluster center in the ResNet-50 feature space.
The comparison results on all 20 classes are reported in Table~\ref{table:supp}.

\begin{figure}[t]
	\centering
	\includegraphics[width=1\linewidth]{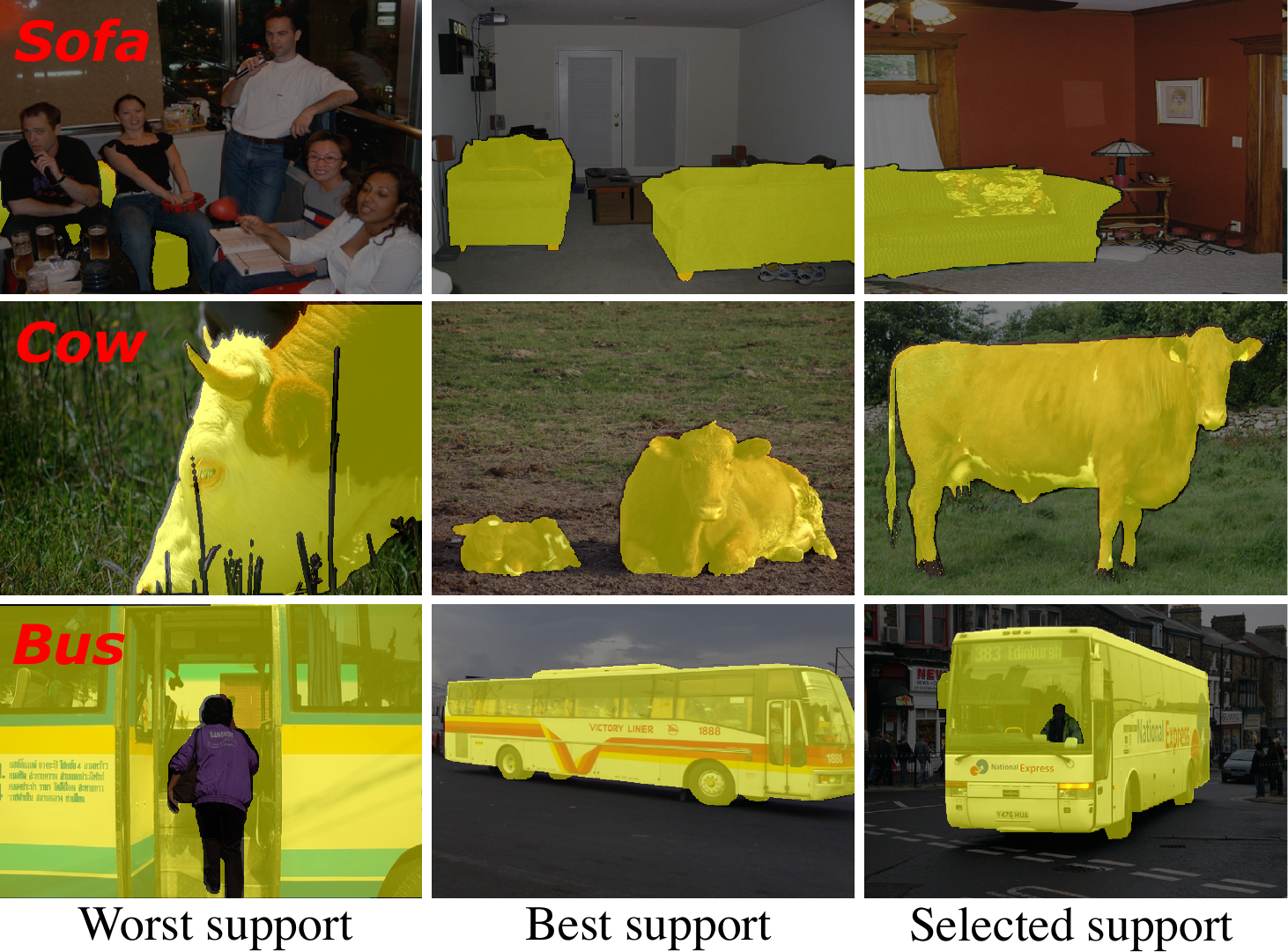}
	\caption{Comparison of the worst, best, and selected support images in three classes.}
	\label{fig:supp}
\end{figure}

From Table~\ref{table:supp}, we can observe that: 
1) Using different images as support can result in very different performance.
Averaged over all 20 classes, the difference between the worst and best IoUs is as large as 49.4\%. 
It indicates that the support image selection is an important factor to consider when using few-shot segmentation in practice.
2) The average IoU of all support images is 4.9\% lower than the IoU of the representative support image (56.3\% vs 61.2\%). 
It means that the using the cluster centers in the deep feature space is a good choice to select support images from a large set of unlabeled images.
In Fig.~\ref{fig:supp}, we display the worst, best and selected support images of three classes. 
Intuitively, a good support image should well represent both the foreground and background classes.

\setlength{\tabcolsep}{1.9mm}
\begin{table}[t]
\centering
\caption{Mean-IoU (\%) results of using different support images on each class of PASCAL-$5^i$. ``Min'' and ``Max'' refer to the performance of the worst and best support images, respectively. ``Avg'' is the performance averaged on all images in the class, and ``Rep'' is the performance of the selected representative image.}
\label{table:supp}
\begin{tabular}{c|l|l|lll|l}
\toprule
\textbf{} & \textbf{Class} & \textbf{Num} & \textbf{Min} & \textbf{Max} & \textbf{Avg} & \textbf{Rep} \\ \midrule
1                  & aeroplane            & 81                 & 54.4                    & 85.7                    & 78.9                     & 85.7                     \\
2                  & bicycle              & 63                 & 15.8                    & 49.8                    & 41.5                     & 47.0                     \\
3                  & bird                 & 89                 & 41.6                    & 84.6                    & 75.0                     & 75.9                     \\
4                  & boat                 & 61                 & 6.9                     & 64.6                    & 47.6                     & 54.0                     \\
5                  & bottle               & 70                 & 2.4                     & 65.3                    & 41.8                     & 35.6                     \\
6                  & bus                  & 70                 & 23.6                    & 85.6                    & 78.5                     & 83.8                     \\
7                  & car                  & 100                & 1.6                     & 68.1                    & 51.8                     & 37.3                     \\
8                  & cat                  & 115                & 19.3                    & 85.6                    & 78.3                     & 82.5                     \\
9                  & chair                & 110                & 2.6                     & 33.1                    & 18.8                     & 30.6                     \\
10                 & cow                  & 69                 & 74.8                    & 89.8                    & 87.4                     & 88.6                     \\
11                 & diningtable          & 73                 & 0.3                     & 44.2                    & 23.9                     & 22.8                     \\
12                 & dog                  & 124                & 14.9                    & 83.2                    & 71.5                     & 80.3                     \\
13                 & horse                & 76                 & 25.8                    & 86.8                    & 81.3                     & 82.4                     \\
14                 & motorbike            & 74                 & 28.7                    & 80.1                    & 70.6                     & 76.7                     \\
15                 & person               & 385                & 0.1                     & 64.5                    & 39.1                     & 52.5                     \\
16                 & pottedplant          & 63                 & 2.3                     & 41.7                    & 26.4                     & 40.7                     \\
17                 & sheep                & 53                 & 48.0                    & 89.5                    & 85.4                     & 88.7                     \\
18                 & sofa                 & 88                 & 5.5                     & 55.0                    & 34.2                     & 38.3                     \\
19                 & train                & 82                 & 23.0                    & 76.7                    & 64. 2                     & 72.5                     \\
20                 & tvmonitor            & 69                 & 1.6                     & 49.0                    & 30.4                     & 48.8                     \\ \midrule
\multicolumn{3}{c|}{Mean} & 19.7                    & 69.1                    & 56.3                     & 61.2     \\
\bottomrule
\end{tabular}
\end{table}

\section{Conclusion}
The domain gap issue seriously limited the application of existing FSS models in domains other than natural images, but was ignored by past literature. 
In this paper, we proposed the first transductive fine-tuning method to address this problem for prototypical FSS models. 
Without ground truth query masks, we employed support labels as implicit supervision to incorporate unlabeled query images into fine-tuning, which is realized by a novel uncertainty-aware semi-supervised loss. 
Our method could simultaneously generalize the base model to the target domain and optimize the segmentation results of the given query set.
Extensive experiments on remote sensing and medical images validated the effectiveness of our methods.

{\small
\bibliographystyle{ieee_fullname}
\bibliography{FSS}

\begin{thebibliography}{10}\itemsep=-1pt

\bibitem{boudiaf2021few}
Malik Boudiaf, Hoel Kervadec, Ziko~Imtiaz Masud, Pablo Piantanida, Ismail
  Ben~Ayed, and Jose Dolz.
\newblock Few-shot segmentation without meta-learning: A good transductive
  inference is all you need?
\newblock In {\em Proceedings of the IEEE/CVF Conference on Computer Vision and
  Pattern Recognition}, pages 13979--13988, 2021.

\bibitem{chen2019closerfewshot}
Wei-Yu Chen, Yen-Cheng Liu, Zsolt Kira, Yu-Chiang Wang, and Jia-Bin Huang.
\newblock A closer look at few-shot classification.
\newblock In {\em International Conference on Learning Representations (ICLR)},
  2019.

\bibitem{chen2019learning}
Xu Chen, Bryan~M Williams, Srinivasa~R Vallabhaneni, Gabriela Czanner, Rachel
  Williams, and Yalin Zheng.
\newblock Learning active contour models for medical image segmentation.
\newblock In {\em Proceedings of the IEEE/CVF Conference on Computer Vision and
  Pattern Recognition}, pages 11632--11640, 2019.

\bibitem{chen2021semi}
Xiaokang Chen, Yuhui Yuan, Gang Zeng, and Jingdong Wang.
\newblock Semi-supervised semantic segmentation with cross pseudo supervision.
\newblock In {\em Proceedings of the IEEE/CVF Conference on Computer Vision and
  Pattern Recognition}, pages 2613--2622, 2021.

\bibitem{dhillon2019baseline}
Guneet~Singh Dhillon, Pratik Chaudhari, Avinash Ravichandran, and Stefano
  Soatto.
\newblock A baseline for few-shot image classification.
\newblock In {\em International Conference on Learning Representations}, 2020.

\bibitem{dong2018few}
Nanqing Dong and Eric~P Xing.
\newblock Few-shot semantic segmentation with prototype learning.
\newblock In {\em British Machine Vision Conference ({BMVC})}, volume~3, 2018.

\bibitem{guo2020broader}
Yunhui Guo, Noel~C Codella, Leonid Karlinsky, James~V Codella, John~R Smith,
  Kate Saenko, Tajana Rosing, and Rogerio Feris.
\newblock A broader study of cross-domain few-shot learning.
\newblock In {\em European Conference on Computer Vision (ECCV)}, pages
  124--141. Springer, 2020.

\bibitem{he2016deep}
Kaiming He, Xiangyu Zhang, Shaoqing Ren, and Jian Sun.
\newblock Deep residual learning for image recognition.
\newblock In {\em Proceedings of the IEEE conference on computer vision and
  pattern recognition}, pages 770--778, 2016.

\bibitem{liu2020crnet}
Weide Liu, Chi Zhang, Guosheng Lin, and Fayao Liu.
\newblock Crnet: Cross-reference networks for few-shot segmentation.
\newblock In {\em IEEE Conference on Computer Vision and Pattern Recognition
  (CVPR)}, pages 4165--4173, 2020.

\bibitem{liu2018learning}
Yanbin Liu, Juho Lee, Minseop Park, Saehoon Kim, Eunho Yang, Sung~Ju Hwang, and
  Yi Yang.
\newblock Learning to propagate labels: Transductive propagation network for
  few-shot learning.
\newblock In {\em International Conference on Learning Representations}, 2019.

\bibitem{liu2020part}
Yongfei Liu, Xiangyi Zhang, Songyang Zhang, and Xuming He.
\newblock Part-aware prototype network for few-shot semantic segmentation.
\newblock In {\em European Conference on Computer Vision (ECCV)}, pages
  142--158. Springer, 2020.

\bibitem{lu2020contour}
Yuhang Lu et~al.
\newblock Contour transformer network for one-shot segmentation of anatomical
  structures.
\newblock {\em IEEE Transactions on Medical Imaging}, 2020.

\bibitem{mittal2019semi}
Sudhanshu Mittal, Maxim Tatarchenko, and Thomas Brox.
\newblock Semi-supervised semantic segmentation with high-and low-level
  consistency.
\newblock {\em IEEE transactions on pattern analysis and machine intelligence},
  2019.

\bibitem{ouali2020semi}
Yassine Ouali, C{\'e}line Hudelot, and Myriam Tami.
\newblock Semi-supervised semantic segmentation with cross-consistency
  training.
\newblock In {\em Proceedings of the IEEE/CVF Conference on Computer Vision and
  Pattern Recognition}, pages 12674--12684, 2020.

\bibitem{paszke2019pytorch}
Adam Paszke, Sam Gross, Francisco Massa, Adam Lerer, James Bradbury, Gregory
  Chanan, Trevor Killeen, Zeming Lin, Natalia Gimelshein, Luca Antiga, et~al.
\newblock Pytorch: An imperative style, high-performance deep learning library.
\newblock {\em Advances in neural information processing systems},
  32:8026--8037, 2019.

\bibitem{russakovsky2015imagenet}
Olga Russakovsky, Jia Deng, Hao Su, Jonathan Krause, Sanjeev Satheesh, Sean Ma,
  Zhiheng Huang, Andrej Karpathy, Aditya Khosla, Michael Bernstein, et~al.
\newblock Imagenet large scale visual recognition challenge.
\newblock {\em International journal of computer vision}, 115(3):211--252,
  2015.

\bibitem{shaban2017one}
Amirreza Shaban, Shray Bansal, Zhen Liu, Irfan Essa, and Byron Boots.
\newblock One-shot learning for semantic segmentation.
\newblock 2017.

\bibitem{scr}
Junji Shiraishi, Shigehiko Katsuragawa, Junpei Ikezoe, Tsuneo Matsumoto,
  Takeshi Kobayashi, Ken-ichi Komatsu, Mitate Matsui, Hiroshi Fujita, Yoshie
  Kodera, and Kunio Doi.
\newblock Development of a digital image database for chest radiographs with
  and without a lung nodule: receiver operating characteristic analysis of
  radiologists' detection of pulmonary nodules.
\newblock {\em American Journal of Roentgenology}, 174(1):71--74, 2000.

\bibitem{snell2017prototypical}
Jake Snell, Kevin Swersky, and Richard~S Zemel.
\newblock Prototypical networks for few-shot learning.
\newblock {\em arXiv preprint arXiv:1703.05175}, 2017.

\bibitem{rooftop}
Xiaolu Sun, C~Mario Christoudias, and Pascal Fua.
\newblock Free-shape polygonal object localization.
\newblock In {\em European Conference on Computer Vision}, pages 317--332.
  Springer, 2014.

\bibitem{sung2018learning}
Flood Sung, Yongxin Yang, Li Zhang, Tao Xiang, Philip~HS Torr, and Timothy~M
  Hospedales.
\newblock Learning to compare: Relation network for few-shot learning.
\newblock In {\em Proceedings of the IEEE conference on computer vision and
  pattern recognition}, pages 1199--1208, 2018.

\bibitem{PFENet}
Z Tian, H Zhao, M Shu, Z Yang, R Li, and J Jia.
\newblock Prior guided feature enrichment network for few-shot segmentation.
\newblock {\em IEEE Transactions on Pattern Analysis and Machine Intelligence},
  2020.

\bibitem{crossdomainfewshot}
Hung-Yu Tseng, Hsin-Ying Lee, Jia-Bin Huang, and Ming-Hsuan Yang.
\newblock Cross-domain few-shot classification via learned feature-wise
  transformation.
\newblock In {\em International Conference on Learning Representations (ICLR)},
  2020.

\bibitem{van2008visualizing}
Laurens Van~der Maaten and Geoffrey Hinton.
\newblock Visualizing data using t-sne.
\newblock {\em Journal of machine learning research}, 9(11), 2008.

\bibitem{wang2019panet}
Kaixin Wang, Jun~Hao Liew, Yingtian Zou, Daquan Zhou, and Jiashi Feng.
\newblock Panet: Few-shot image semantic segmentation with prototype alignment.
\newblock In {\em IEEE International Conference on Computer Vision (ICCV)},
  pages 9197--9206, 2019.

\bibitem{yang2021mining}
Lihe Yang, Wei Zhuo, Lei Qi, Yinghuan Shi, and Yang Gao.
\newblock Mining latent classes for few-shot segmentation.
\newblock {\em arXiv preprint arXiv:2103.15402}, 2021.

\bibitem{zhang2019canet}
Chi Zhang, Guosheng Lin, Fayao Liu, Rui Yao, and Chunhua Shen.
\newblock Canet: Class-agnostic segmentation networks with iterative refinement
  and attentive few-shot learning.
\newblock In {\em IEEE Conference on Computer Vision and Pattern Recognition
  (CVPR)}, pages 5217--5226, 2019.

\bibitem{zou2020pseudoseg}
Yuliang Zou, Zizhao Zhang, Han Zhang, Chun-Liang Li, Xiao Bian, Jia-Bin Huang,
  and Tomas Pfister.
\newblock Pseudoseg: Designing pseudo labels for semantic segmentation.
\newblock {\em arXiv preprint arXiv:2010.09713}, 2020.

\end{thebibliography}
}

\end{document}